\pdfoutput=1

\documentclass[11pt]{article}

\usepackage[preprint]{acl}

\usepackage{times}
\usepackage{latexsym}

\usepackage[T1]{fontenc}

\usepackage[utf8]{inputenc}

\usepackage{microtype}

\usepackage{inconsolata}

\usepackage{graphicx}
\usepackage{arydshln} 
\usepackage{booktabs}
\usepackage{multirow}
\usepackage{adjustbox}
\usepackage{enumitem}
\usepackage{amsmath}
\usepackage{amssymb}
\usepackage{pifont}
\usepackage{makecell}
\usepackage{footmisc}
\usepackage{booktabs}
\usepackage{tabularx} 
\usepackage{CJKutf8}
\usepackage{hyperref}
\usepackage{subcaption}
\usepackage{url}

\title{Zero-Shot Conversational Stance Detection: Dataset and Approaches}

\author{Yuzhe Ding, \ Kang He, \ Bobo Li, \  Li Zheng,  \ Haijun He, \ Fei Li, \\ 
\ \textbf{Chong Teng,} \ \textbf{Donghong Ji\thanks{
Corresponding author.}} \\
  Key Laboratory of Aerospace Information Security and Trusted Computing, Ministry of \\ Education, School of Cyber Science and Engineering, Wuhan University \\
  \texttt{\{yuzheding,lifei\_csnlp,dhji\}@whu.edu.cn} \\
}

\begin{document}
\maketitle
\begin{abstract}
Stance detection, which aims to identify public opinion towards specific targets using social media data, is an important yet challenging task.
With the increasing number of online debates among social media users, conversational stance detection has become a crucial research area.
However, existing conversational stance detection datasets are restricted to a limited set of specific targets, which constrains the effectiveness of stance detection models when encountering a large number of unseen targets in real-world applications.
To bridge this gap, we manually curate a large-scale, high-quality zero-shot conversational stance detection dataset, named ZS-CSD, comprising 280 targets across two distinct target types.
Leveraging the ZS-CSD dataset, we propose SITPCL, a speaker interaction and target-aware prototypical contrastive learning model, and establish the benchmark performance in the zero-shot setting.
Experimental results demonstrate that our proposed SITPCL model achieves state-of-the-art performance in zero-shot conversational stance detection.  
Notably, the SITPCL model attains only an F1-macro score of 43.81\%, highlighting the persistent challenges in zero-shot conversational stance detection. Our data and code are available at \href{https://github.com/whu-yzding/ZS-CSD}{https://github.com/whu-yzding/ZS-CSD}.
\end{abstract}

\section{Introduction}

With the rise of social networks, people increasingly express their opinions views online.
The aim of stance detection is to determine people’s opinionated standpoint or attitude (e.g. Favor, Against, or Neutral ) expressed in text towards a specific target \citep{somasundaran-wiebe-2010-recognizing,mohammad-etal-2016-semeval,sobhani-etal-2017-dataset,aldayel2021stance}, such as the COVID-19 vaccine, LGBTQ rights, and educational philosophy.
Stance detection has a wide range of potential applications, including in fields such as emotion recognition \citep{zhang-etal-2020-enhancing-cross,ijcai2023p693,zheng2023ecqed}, argument mining \citep{sobhani-etal-2015-argumentation,rajendran-etal-2018-something,sirrianni-etal-2020-agreement,10.1145/3539618.3591917,zheng2025enhancing}, and rumor detection \citep{yu-etal-2020-coupled,yang-etal-2024-reinforcement,zheng2025multi}.

Existing research on stance detection is typically divided into in-target, cross-target, and zero-shot categories for training and evaluating models, with a primary focus on analyzing individual sentences \citep{zarrella-marsh-2016-mitre,kuccuk2020stance,allaway-mckeown-2020-zero,li-etal-2023-new}.
However, on social media platforms, users not only express opinions through direct posts but also engage in conversations and express their views in the comment sections of tweets they find interesting. Multi-party conversation understanding involves participant diversity, interaction dynamics, context dependence, and rich emotions and tones, making stance detection in conversational scenarios particularly challenging.

Currently, five conversational stance detection datasets have been developed, including SRQ \citep{villa2020stance}, CANT-CSD \citep{li2023improved}, CTSDT \citep{li2023contextual}, MT-CSD \citep{niu-etal-2024-challenge} and MmMtCSD \citep{niu2024multimodal}. 
Despite the diversity of conversational stance detection corpora, these datasets have several limitations. 
Firstly, current conversational stance detection datasets focus only on a few specific topics and only contain noun phrase type targets or the post type targets. The number of targets is relatively small, and the largest MT-CSD contains only five targets.
Secondly, in the previous conversation stance detection dataset annotation and modeling process, only the reply relationship and historical context information were considered during the annotation and modeling process, without considering the speaker context information and potential speaker interaction information. 
Finally, current research in conversational stance detection is restricted to two types of tasks: in-target and cross-target stance detection, lacking exploration of zero-shot conversational stance detection tasks that better reflect real-world scenarios.

To address these issues, we introduce ZS-CSD, the first zero-shot multi-turn, multi-party conversational stance detection dataset. This dataset encompasses a substantial number of real-world human interactions on trending topics, events, and opinions from the Chinese online social media platform Weibo.
Compared to existing conversational stance detection datasets, the ZS-CSD dataset captures a more diverse range of targets, including noun phrase type targets (i.e. entities, events, or topics) and claim type targets (i.e. phrases that express opinions). Furthermore, the stance annotation process in ZS-CSD thoroughly considers speaker context information and potential speaker interaction information, aspects that have been overlooked in previous studies. The specific details of ZS-CSD and other stance detection datasets can be seen in Table \ref{Table 1}.

Based on the ZS-CSD dataset, we design a new conversational stance detection task: \textit{target-based zero-shot conversational stance detection}, where a number of completely unseen targets are used to evaluate the conversational stance detection model.
Our new task challenges in two main aspects.
First, the model needs to learn how to judge the user's stance toward a given target from the training set and transfer the ability to unseen targets in the test set.
Second, the model needs to fully understand the speaker context information and potential speaker interaction information in the conversation.

To solve these challenges, we present a SITPCL model. 
Specifically, we construct a speaker interaction network to model intra-speaker and inter-speaker dependencies. 
Additionally, we introduce a target-aware prototypical contrastive learning method to enable the model to better understand the targets.
Experimental results demonstrate that our proposed SITPCL model achieves state-of-the-art performance in zero-shot conversational stance detection.

The main contributions of this paper can be summarized as follows:
\begin{itemize}[label={\scriptsize$\bullet$},leftmargin=1em]
    \item 
    We propose the first zero-shot multi-turn multi-party conversational stance detection dataset ZS-CSD, and propose a new target-based zero-shot conversational stance detection task.
    \item 
    Compared to previous conversational stance detection datasets, our ZS-CSD incorporates a wider range of target types and an increased number of targets. Additionally, we take into account speaker context information and speaker interaction information. The final ZS-CSD dataset comprises 17,063 conversation samples from 8,667 users, including 113 noun phrase targets and 167 claim targets.
    \item
    We introduce a speaker interaction and target-aware prototypical contrastive learning model SITPCL to benchmark the ZS-CSD task. Our model effectively learns intra-speaker and inter-speaker dependencies and can better understand the unseen targets in the zero-shot setting.
\end{itemize}

\section{Related Work}
\subsection{Stance Detection Datasets}
\begin{table*}[t]
\centering
\resizebox{1\linewidth}{!}{
\begin{tabular}{llccccccc}
\toprule
\multirow{2}{*}{\textbf{Name}} & \multirow{2}{*}{\textbf{Source}} & \multirow{2}{*}{\textbf{Type}} & \multirow{2}{*}{\textbf{Task}}                                            & \multicolumn{1}{l}{\multirow{2}{*}{\textbf{\# Target(s)}}} & \multicolumn{2}{c}{\textbf{Target Type}} & \multirow{2}{*}{\textbf{Language}} & \multicolumn{1}{l}{\multirow{2}{*}{\textbf{Size}}} \\ \cline{6-7}
&              &          &                                                                  & \multicolumn{1}{l}{}                             & \textbf{NP}        & \textbf{C}       &       & \multicolumn{1}{l}{} \\ 
\midrule
SEM16 & X & S & In-target & 6 &  $\checkmark$ & $\times$ & English & 4,870 \\ 
P-stance  & X & S& In-target & 3 & $\checkmark$ & $\times$ & English & 21,574 \\ 
WT-WT & X  & S & In-target & 5 & $\checkmark$ & $\times$ & English  & 51,284  \\ 
COVID19  & X & S & In-target & 4 & $\checkmark$ & $\times$ & English & 6,133 \\ 
\midrule
VAST & Comments& S & ZSSD  & 5,634                 & $\checkmark$ & $\times$ & English  & 18,545 \\ 
C-STANCE & Weibo  & S & ZSSD & 40,204              & $\checkmark$ & $\checkmark$ & Chinese & 48,126 \\ 
EZ-STANCE & X & S & ZSSD & 40,678           & $\checkmark$ & $\checkmark$ & English & 47,316 \\ 
\midrule
SRQ & X & C & In-target & 4                  & $\checkmark$ & $\times$ & English & 5,220 \\ 
CANT-CSD  & HK SM & C & In-target & 1    & $\checkmark$ & $\times$ & Cantonese & 5,876  \\ 
CTSDT & X & C & In-target & 1                & $\checkmark$ & $\times$ & English  & 53,861 \\ 
STANCEOSAURUS & X & C & Cross-target & 251                & $\times$ & $\checkmark$ & English,Hindi,Arabic  & 28,033 \\
MT-CSD & Reddit & C & \makecell{In-target\\Cross-target}  & 5  & $\checkmark$ & $\times$ & English  & 15,876 \\ 
MmMtCSD & Reddit & C & \makecell{In-target\\Cross-target}  & 3  & $\checkmark$ & $\times$ & English  & 21,340 \\ 
\midrule
ZS-CSD(Ours) & Weibo & C & ZS-CSD & 280         & $\checkmark$ & $\checkmark$ & Chinese & 17,063 \\ 
\bottomrule
\end{tabular}
}
\caption{Comparison of stance detection datasets: Type S denotes sentence-level data, while Type C indicates conversation-level data. Target Type NP refers to noun phrase targets, and Target Type C refers to claim targets.}
\label{Table 1}
\end{table*}

\paragraph{Sentence-Level stance detection datasets}
Over the years, various datasets have been introduced as benchmarks for sentence-level stance detection, as summarized in Table \ref{Table 1}.
The SEM16 dataset \citep{mohammad-etal-2016-semeval} pioneered stance detection from X, while P-Stance \citep{li-etal-2021-p} focuses on longer political tweets.
The large-scale WT-WT corpus \citep{conforti-etal-2020-will} offers extensive labeled data, and specialized datasets like the COVID-19 Stance Detection dataset \citep{glandt-etal-2021-stance} target specific discourse.
The VAST dataset \citep{allaway-mckeown-2020-zero} is the most comprehensive, enabling zero-shot stance detection across thousands of topics. More recently, C-STANCE \citep{zhao-etal-2023-c} and EZ-STANCE \citep{zhao-caragea-2024-ez} have expanded zero-shot stance detection using Weibo and X data.

\paragraph{Conversation-level stance detection datasets}
In real-world scenarios, users express perspectives conversationally, making conversational stance detection—identifying stances within discussion threads—increasingly relevant. The SRQ dataset \citep{villa2020stance} pioneered stance detection in comment data but was limited to single-turn replies and shallow conversations. CANT-CSD \citep{li2023improved} addressed multi-turn interactions with a deeper commenting corpus, while CTSDT \citep{li2023contextual} relied primarily on automated annotations. Stanceosaurus \cite{zheng-etal-2022-stanceosaurus} constructed a conversational stance classification dataset focused on misinformation-related claims. The dataset is designed to facilitate the identification of misinformation by analyzing the stances expressed within the surrounding discussion. MT-CSD \citep{niu-etal-2024-challenge} expanded conversation depth across five targets, and MmMtCSD \citep{niu2024multimodal} extended it to multimodal scenarios for broader applicability.

\subsection{Stance Detection Approaches}
In recent years, various approaches based on traditional machine learning and deep learning have been proposed to address stance detection for specific targets \citep{augenstein-etal-2016-stance,10.1145/3331184.3331367}. The task settings can generally be categorized into in-target, cross-target, and zero-shot settings. Most previous work has focused on in-target stance detection, where a classifier is trained and evaluated on the same target \citep{mohammad-etal-2016-semeval, li-caragea-2019-multi, li-caragea-2021-target}. However, it is often challenging to obtain sufficient annotated data for each target, and conventional models tend to perform poorly when generalized to unseen targets. This limitation has motivated research on cross-target stance detection \citep{xu-etal-2018-cross, zhang-etal-2020-enhancing-cross, li-etal-2021-improving-stance}, where a classifier is adapted to different but related targets. Furthermore, zero-shot stance detection (ZSSD), which aims to identify stances toward a large number of unseen targets, has garnered considerable attention in recent years \citep{liu-etal-2021-enhancing, liang-etal-2022-jointcl, luo-etal-2022-exploiting}.

All current methods for stance detection can generally be classified into fine-tuning based approaches \citep{10.1145/3477495.3531807, liang-etal-2024-multi, wang-pan-2024-target} and LLM-based approaches \citep{li-etal-2023-stance, weinzierl-harabagiu-2024-tree, zhang-etal-2024-llm-driven}. Fine-tuning based methods involve adding a fully connected layer to the [CLS] token of a pre-trained model and fine-tuning the model specifically for stance detection. Besides, large language models (LLMs) have demonstrated remarkable capabilities across diverse applications due to their inherent semantic understanding \citep{zhang2023investigatingchainofthoughtchatgptstance, li2024mitigating,zheng2024reverse}. This semantic understanding presents exciting opportunities for stance detection. Most LLMs can effectively perform stance prediction via zero-shot prompting, significantly enhancing usability \citep{ding-etal-2024-edda}.

\section{Dataset Construction}
\paragraph{Data Collection}
To obtain authentic multi-party conversations with diverse targets, we collected the original corpus of our ZS-CSD dataset from Weibo\footnote{https://weibo.com/}, China's largest social media platform. 
We utilized the Weibo API\footnote{https://open.weibo.com/} to collect 2 million posts along with the online discussions under each post from 80 verified and widely followed news bloggers.
After obtaining the raw data, we selected a series of keywords representing controversial topics for data screening and classified them into six domains.
The list of domains and keywords are provided in Appendix \ref{appendix:A}.
To ensure the high quality of the ZS-CSD dataset, we implemented two rigorous preprocessing steps.
Firstly, we observed that certain comments, which directly respond to the original post, can incite heated discussions. To address this, we selected all first-level comments that directly reply to the original post as the root node of the discussion tree. Subsequently, we gradually constructed multiple discussion tree as candidates based on the reply relationship.
Secondly, to balance the number of conversation samples across different depths in the dataset, we collected samples from the discussion tree based on their depth. Ultimately, we obtained a diverse set of samples spanning six levels \(1, 2, 3, 4, 5, \geq 6\) for subsequent annotation.

\begin{table}[t]
\centering
\resizebox{1\linewidth}{!}{%
\begin{tabular}{ccccccccccccc}
\toprule 
\multirow{2}{*}{}   & \multicolumn{2}{c}{\# Targets}   & \multicolumn{2}{c}{\# Conversations}    & \multirow{2}{*}{Avg. Length} \\  \cmidrule(lr){2-3} \cmidrule(lr){4-5} 
                                   & N              & C               & N               & C                                \\  
\midrule
\rule{0pt}{10pt} Train             & 79             & 111             & 5,604            & 6,299                       & 37.84       \\  
\rule{0pt}{10pt} Val               & 17             & 28              & 961             & 1615                       & 37.28       \\  
\rule{0pt}{10pt} Test              & 17             & 28              & 967             & 1,617                        & 36.98       \\  
\midrule
\rule{0pt}{10pt} Overall           & 113            & 167             & 7,532            & 9,531                     & 37.63      \\
\bottomrule
\end{tabular}%
}
\caption{The dataset split statistics for the ZS-CSD are presented, where N and C denote noun phrase type targets and claim type targets, respectively. Avg. Length refers to the average number of characters per utterance.}
\label{Table 3}
\end{table}

\paragraph{Human Annotation}
We built our annotation platform using Doccano\footnote{https://github.com/doccano/doccano}, with each thread sample as a separate annotation task. Three expert annotators and five NLP researchers, all native Chinese speakers, participated in the process. Annotation was conducted in two steps: target annotation followed by stance annotation.
In Step 1, three expert annotators collaboratively identify one or two targets per conversation after thoroughly reviewing the discussion. The selected targets must: (i) allow for clear stance expression, (ii) avoid near-unanimous stances, such as those violating public values, and (iii) be either noun phrase targets (e.g., entities, concepts, events) or claim targets (e.g., opinions). This process yielded 17,063 conversation-target pairs for stance annotation in Step 2.
In step 2, given 17,063 pairs, five annotators were asked to combine conversation history, speaker context, and speaker interaction information to annotate the current utterance’s stance toward the given target: \textit{favor}, \textit{against}, or \textit{neutral}.
Neutrality to the target can occur in two cases:
i) The user neither supports nor opposes the target. 
ii) The comment is irrelevant to the target.
To ensure data quality and reliability, each data instance is annotated by at least two annotators. In cases of disagreement between the annotators, a panel of at least two expert annotators will discuss the issue to reach a final judgment.
After obtaining the annotation results, we calculated a Cohen’s Kappa score of 0.83, indicating that our annotated corpus is of high quality.


\begin{table}[t]
    \centering
    \resizebox{1\linewidth}{!}{%
    \begin{tabular}{ccccccc}
    \toprule
     & \multicolumn{6}{c}{Depth of Conversation}  \\ 
    \cmidrule(lr){2-7} & 1 & 2 & 3 & 4 & 5 & >=6 \\
    \midrule
    \rule{0pt}{10pt} Train & 210 & 1,905 & 3,732 & 2,548 & 1,636  & 1,872 \\ 
    \rule{0pt}{10pt} Val & 45 & 399 & 832 & 530 & 345 & 425 \\ 
    \rule{0pt}{10pt} Test & 45 & 445 & 877 & 585 & 351 & 281 \\ 
    \midrule
    \rule{0pt}{10pt} Overall & 300 & 2,749 & 5,441 & 3,663 & 2,332 & 2,578 \\
    \bottomrule
    \end{tabular}%
    }
    \caption{Statistics of samples at various depths in the ZS-CSD dataset are presented.}
    \label{Table 4}
\end{table}

\paragraph{Dataset Statistics}
Following the VAST dataset guidelines \citep{allaway-mckeown-2020-zero}, we split the annotated data into training(70\%), validation(15\%), and test(15\%) sets for target-based zero-shot conversational stance detection. Each set contains distinct conversations and targets with no overlap. The dataset distribution is detailed in Table \ref{Table 3} and Table \ref{Table 4}.

\section{Our Approach}

\begin{figure*}[t]
  \centering
  \includegraphics[width=0.9\linewidth]{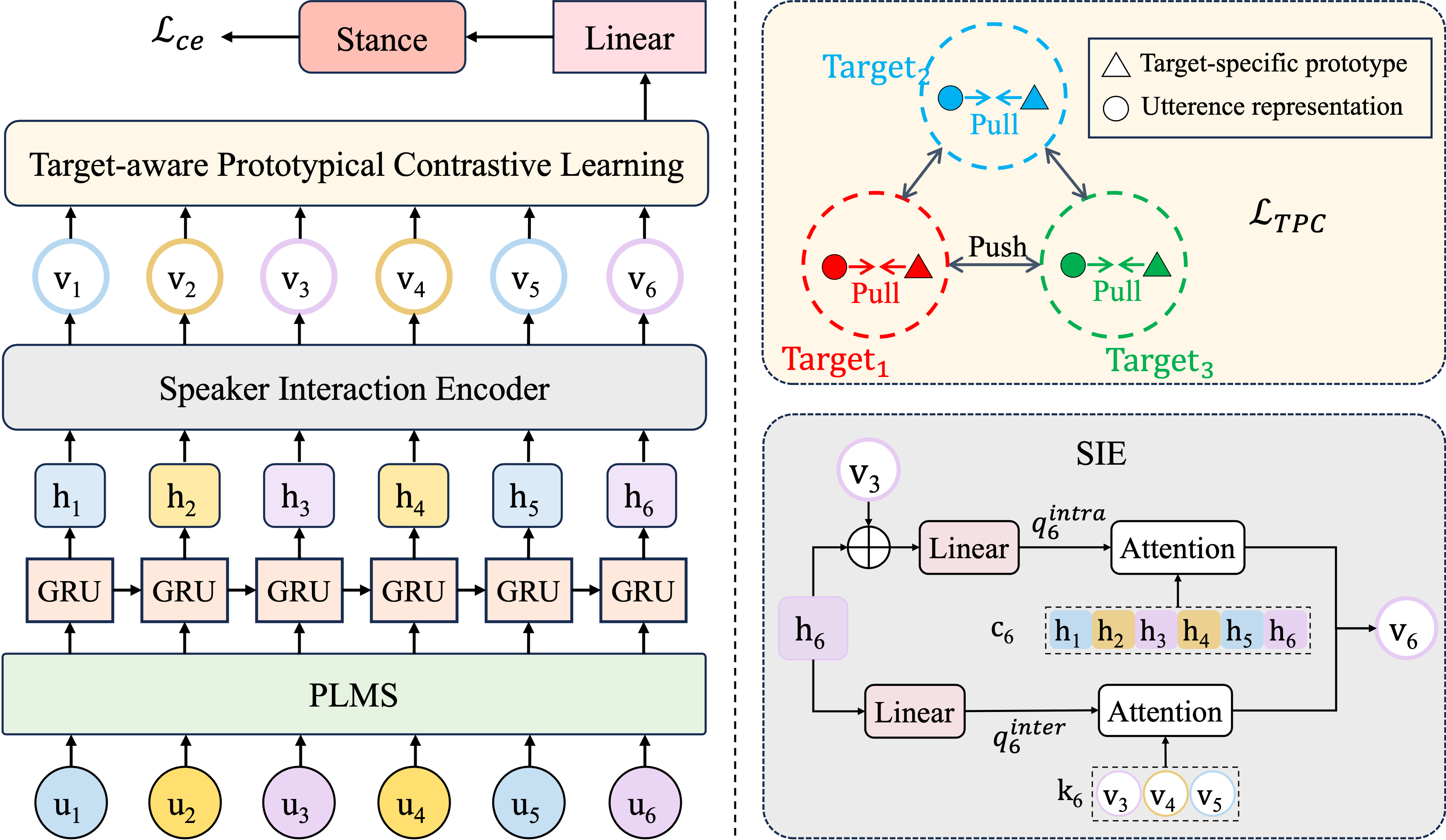}
  \caption{The architecture of our SITPCL framework. Initially, we use the pre-trained language model and GRU network to obtain a contextualized representation of the entire conversation. Next, we capture speaker interactions using the Speaker Interaction Encoder. To enhance representation learning, we incorporate target-aware prototypical contrastive learning. Finally, the enriched utterance representation is leveraged for stance detection.} 
  \label{fig:2}
\end{figure*}

\subsection{Task Definition}
The primary objective of the zero-shot conversational stance detection task is to determine the stance of the current utterance toward a given target, utilizing both the conversation history and speaker context. Notably, the targets in the test set are unseen in both the training and validation sets. Formally, given a target $t$, a multi-turn conversation $C$, a collection of speakers $S$ and a set of stance labels $\mathcal{E}$, if conversation $C$ consists of $n$ utterances, it can be denoted as $C={(s_i,u_i)}_{i=1}^{n}$, where $u_i$ is the utterance of $i$-th turn and $s_i\in S$ is the corresponding speaker of $u_i$. If we denote $u_n$ as the current utterance, the model can only take $(s_n,u_n,t)$ and the previous utterances $\{({s_i,u_i})\}_{i=1}^{n-1}$ as input to predict the stance label $y_n$ of $u_n$ toword $t$, where $y_n \in \mathcal{E}$.

\subsection{Utterance Encoder}
We adopt a pre-trained language model (PLM), such as BERT \citep{devlin-etal-2019-bert}, to encode utterance-target pairs within the conversation. Since the entire conversation may exceed BERT's maximum input length, we encode each utterance-target pair individually using separate PLMs.
We concatenate each utterance $u_i$ with its speaker $s_i$ and add a template with a placeholder to implicitly express the speaker’s stance:
\begin{equation}
m_i = PLM([s_i, \mathbf{u}_i, \langle /s \rangle, \textit{template}, \langle /s \rangle])
\end{equation}
\vspace{-1em} 
\begin{equation}
\textit{template} = \ s_i \ \text{expresses} \ \langle \text{mask} \rangle \ \text{towords} \ \text{target}
\end{equation}
where $</s>$ denotes the separator used to distinguish different contents, $<mask>$ is the placeholder (known as the masked token) used for learning the stance label toward given target expressed in this utterance.
To get the contextual representations, {$m_1,m_2,...,m_n$} are passed to a GRU layer \citep{cho-etal-2014-learning}. The final representations are defined by:
\begin{equation}
[h_0, h_1, \ldots, h_n] = \text{GRU}([m_0, m_1, \ldots, m_n])
\end{equation}
where \( h_i \in \mathbb{R}^d \). Here, \( d \) is the dimension of each utterance vector.

\subsection{Speaker Interaction Encoder}
To accurately capture the dynamics of speaker interactions, we propose an encoder network specifically designed to model both inter-speaker and intra-speaker dependencies.

\paragraph{\textbf{Intra-Speaker Dependency Modeling.}}
For a given utterance $u_i$ which is uttered by speaker $s_{u_i}$, we take the last previous speaker enhanced representation of $s_{u_i}$ and previous context $c_i$ = $\{h_1, h_2, ..., h_i\}$ as input, modeling the intra-speaker dependency implied in the conversation.

First, we utilize the last previous speaker enhanced representation of $s_{u_i}$ and the representation vector of $u_i$ to obtain the query vector $q_i^{intra}$ as follows:
\begin{equation}
{q}_i^{\text{intra}} = W_q^{\text{intra}} \left[ v_{s_{u_i}} \oplus h_i \right] + b_q^{\text{intra}}
\end{equation}
where $v_{s_{(u_i)}} \in \mathbb{R}^{h \times 1}$ means the last previous speaker enhanced representation of $s_{u_i}$; $\oplus$ means the concatenation operation; $W_q^{\text{intra}} \in \mathbb{R}^{h \times 2h}$ and $b_q^{\text{intra}} \in \mathbb{R}^{1}$ are model parameters.

An attention mechanism is then introduced to model the intra-speaker context, with the query vector $q_i^{intra}$ acting as the query and the previous context $c_i$ serving as the key and value in the attention framework. This mechanism generates the intra-speaker state vector $v_i^{intra}$ as follows:
\begin{equation}
\alpha_i^{\text{intra}} = \text{softmax}\left( \mathbf{W}_1 \left( \mathbf{q}_i^{\text{intra}} \odot \mathbf{c}_i \right) + \mathbf{b}_1 \right)
\end{equation}
\vspace{-1em} 
\begin{equation}
\mathbf{v}_i^{\text{intra}} = \alpha_i^{\text{intra}} \ \circ \ \mathbf{c}_i^T
\end{equation}

\paragraph{\textbf{Inter-Speaker Dependency Modeling.}}
To model inter-speaker dependencies, we use the context vector $h_i$. The key vector $k_i$ for the attention mechanism is calculated prior local information $\{v_j \ | \ j < i\}$ linked to speaker $s_{u_i}$. This approach captures the latent inter-speaker dependency. To derive the inter-speaker state vector for $u_i$, we apply the following procedure:
\begin{equation}
q_i^{\text{inter}} = W_q^{\text{inter}} h_i + b_q^{\text{inter}}
\end{equation}
\vspace{-1em} 
\begin{equation}
\alpha_i^{\text{inter}} = \text{softmax}\left( \mathbf{W}_2 \left( \mathbf{q}_i^{\text{inter}} \ \odot \ \mathbf{k}_i \right) + \mathbf{b}_2 \right)
\end{equation}
\vspace{-1em} 
\begin{equation}
\mathbf{v}_i^{\text{inter}} = \alpha_i^{\text{inter}} \ \circ \ \mathbf{k}_i^T
\end{equation}

After obtaining the intra-speaker and inter-speaker state vectors, $v_i^{intra}$ and $v_i^{inter}$ , we combine them to obtain the current speaker enhanced representation $v_i$, where $W_3 \in \mathbb{R}^{h \times 2h}$ and $b_3 \in \mathbb{R}^{1}$ are model parameters.
\begin{equation}
v_i = W_3 \left[v_i^{\text{intra}} \oplus v_i^{\text{inter}} \right] + b_3
\end{equation}

\subsection{Target-aware Prototypical Contrastive Learning}
Contrastive learning aims to learn an effective representation space by making samples closer to their semantically similar positive samples and away from other samples. Based on this, we design a target-aware prototype-level contrastive learning objective in a mini-batch. 

Taking target $t$ as an example, its prototype representation $p_t$ is defined as the average of all utterances representation associated with $t$ . The $p_t$ is iteratively refined as the corresponding utterances representation are optimized. The target-aware prototypical contrastive loss is:

\begin{equation}
p_t = \frac{1}{|C_t|} \sum_{i \in C_t} \mathbf{v}_i
\end{equation}
\vspace{-1em} 
\begin{equation}
\mathcal{L_{\text{TPC}}} = -\frac{1}{N} \sum_{i=1}^{N} \log \frac{\exp(\text{sim}(\mathbf{v}_i, p_{y_i}) / \tau)}{\sum_{k=1}^{K} \exp(\text{sim}(\mathbf{x}_i, p_k) / \tau)}
\end{equation}
where $C_t$ is the number of utterances for the target $t$ in a mini-batch, $p_t$ is the prototype representation of the target $t$, $sim(.,.)$ is the similarity metric, $\tau$ is the temperature parameter, $K$ is the number of targets, and $N$ is the number of all utterances.

Therefore, the representation of any utterance in the mini-batch will move closer to its corresponding target prototype while distancing itself from the prototype representations of other targets. Simultaneously, the prototype representations of different targets will progressively diverge from one another.

\subsection{Stance Detection}
The final output representation of the current utterance $v_i$ is then input into a classifier with a softmax function to produce a predicted stance distribution $\hat{y}_i \in \mathbb{R}^{d_y}$.
\begin{equation}
  \label{eq:example}
    \hat{y}_i = \text{softmax}(W \cdot v_i + b)
\end{equation}
where \(d_y\) is the dimensionality of stance labels. 
\(W \in \mathbb{R}^{d_y \times h}\) and \(b \in \mathbb{R}^{d_y}\) are trainable parameters. 
We adopt a cross-entropy loss between predicted distribution \(\hat{y}_i\) and ground-truth distribution \(y_i\) to train the classifier:
\begin{equation}
\mathcal{L}_{\text{CE}} = - \sum_{i=1}^{N_b} \sum_{j=1}^{d_y} y_i^j \log \hat{y}_i^j
\end{equation}

\subsection{Learning Objective}
The learning objective of our proposed model is to train the model by jointly optimizing a supervised loss of stance detection $\mathcal{L}_{\text{CE}}$ with a target-aware prototypical contrastive loss. The overall loss $\mathcal{L}$ is formulated by summing up two losses together:
\begin{equation}
\mathcal{L} = \mathcal{L}_{\text{CE}} + \gamma \mathcal{L}_{\text{TPC}}
\end{equation}
where $\gamma$ are the hyper-parameter for weights balance.

\section{Experimental Setup}

\begin{table*}[t]
\centering
\resizebox{1\linewidth}{!}{%
\renewcommand{\arraystretch}{1.3}
\begin{tabular}{lcccccccccccc}
\toprule
 \multirow{2}{*}{\textbf{METHOD}} & \multicolumn{4}{c}{\textbf{Mixed targets}} & \multicolumn{4}{c}{\textbf{Noun-phrase targets}} & \multicolumn{4}{c}{\textbf{Claim targets}} \\  \cmidrule(lr){2-5} \cmidrule(lr){6-9} \cmidrule(lr){10-13}
                         & Favor & Against & Neutral & All   & Favor   & Against   & Neutral  & All    & Favor & Against & Neutral & All   \\ 
\midrule
 Llama3-8B & 35.09 & 39.34   & 33.29   & 35.91 & 38.31   & 42.68     & 37.13    & 39.37  & 33.26 & \underline{37.44}   & 30.68   & 33.79 \\ 
 GPT4o-mini & 24.75 & 38.63   & \underline{51.10}  & 38.16 & 20.78 & 46.86 & 52.99 & 40.21 & 26.87 & 32.77 & \underline{50.07} & 36.57 \\ 
 GPT3.5 & 28.04 & 38.89   & \textbf{53.84}  & 40.25 & 38.41 & 48.79 & \underline{54.84} & 47.35 & 20.99 & 32.29 & \textbf{53.32} & 35.53 \\ 
 Llama3-70B & 40.81 & 38.04  & 44.37  & 41.07 & \textbf{48.77}  & \underline{50.40}  & 47.00  & \textbf{48.72} & 35.96 & 30.98  & 42.75  & 36.56 \\ 
 Qwen2.5-14B & 33.84 & \textbf{44.01}  & 50.48  & \underline{42.78} & 27.67  & \textbf{50.56}  & \textbf{55.58}  & 44.60 & 36.86 & \textbf{40.23}  & 46.93  & \underline{41.34} \\
\cdashline{1-13}
 Roberta  & 37.37 & 38.94   & 44.50   & 40.27 & 37.50   & 42.57    & 48.31    & 42.79  & 37.30 & 36.64   & 42.20   & 38.71 \\
 Branch-BERT  & \underline{43.11} & 38.19   & 41.97   & 41.09 & \underline{46.29}  & 43.73     & 43.41    & 44.48  & \underline{41.43} & 34.34   & 41.13   & 38.96 \\
 GLAN  & \textbf{43.68} & 34.91  & 46.79  & 41.79 & 45.91  & 37.65   & 51.58    & 45.04  & \textbf{42.43} & 33.18  & 43.88  & 39.83 \\ 
  \textbf{SITPCL(Ours)} & 41.03 & \underline{41.34} & 49.07  & \textbf{43.81} & 40.99 & 47.14 & 54.49 & \underline{47.54} & 41.06 & 37.34  & 46.02 & \textbf{41.47} \\ 
\bottomrule
\end{tabular}%
}
\caption{We compare different models on the target-based zero-shot conversational stance detection task. The performance is evaluated using the F1 score for the categories: favor, against, neutral, and the overall F1-macro. }
\label{Table 5}
\end{table*}

\begin{table}[t]
\centering
\resizebox{1\linewidth}{!}{%
\renewcommand{\arraystretch}{1.2}
\begin{tabular}{lcccc}
\toprule
 \multirow{2}{*}{\textbf{METHOD}} & \multicolumn{4}{c}{\textbf{Mixed targets}}  \\  \cmidrule(lr){2-5} 
                         & Favor & Against & Neutral & All     \\ 
\midrule
  \textbf{SITPCL(Ours)} & 41.03 & 41.34 & 49.07  & 43.81  \\ 
 \quad W/o SIE & 40.11 & 43.46 & 45.15  & 42.91  \\ 
 \quad W/o TPCL & 31.50 & 47.26 & 49.02  & 42.59  \\
 \quad W/o BOTH & 37.80 & 42.14 & 44.28  & 41.40  \\
\bottomrule
\end{tabular}%
}
\caption{Experimental results of ablation study.}
\label{Table 6}
\end{table}

\subsection{Baseline Methods}

\paragraph{Fine-tuning with PLMs.} 
(1) RoBERTa \citep{liu2019roberta} represents an enhancement over BERT, utilizing larger batch sizes and more data for training;
(2) Branch-BERT \citep{li2023improved} utilizes a CNN to extract important ngrams features incorporating contextual information in conversation threads.
(3) GLAN \citep{niu-etal-2024-challenge} architecture adopts a three-branch structure to address the intricacies of conversational dynamics comprehensively.

\paragraph{LLM-based methods.} 
We use current state-of-the-art LLMs with various parameter sizes, including the closed-source GPT3.5\footnote{https://platform.openai.com/docs/models/gpt-3-5}, GPT4o-mini\footnote{https://platform.openai.com/docs/models/gpt-4o-mini} and open-source LLMs such as Qwen2.5-14B\footnote{https://huggingface.co/Qwen/Qwen2.5-14B-Instruct}, Llama3-70B\footnote{https://huggingface.co/meta-llama/Meta-Llama-3-70B-Instruct} and Llama3-8B\footnote{https://huggingface.co/meta-llama/Meta-Llama-3-8B-Instruct}. We evaluate these models exclusively under the zero-shot prompt setting. The prompt details are provided in Appendix \ref{appendix:B}.

\subsection{Training Settings and Evaluation Metrics}
We perform experiments using two NVIDIA RTX 3090 GPUs. We initialize all the pre-trained language models using Chinese-Roberta-wwm-ext \citep{9599397}. The hidden dimension of the GRU layer is set to 768 to match the output dimension of the PLM. During training, we use a batch size of 16 and run for 20 epochs. The total training time is less than 3 hours. The optimizer is AdamW, with a learning rate of 1e-5 along with a weight decay of 1e-6. We employ five different random seeds for all experiments, and the final score is the average of these five runs. Consistent with previous studies \citep{zhao-etal-2023-c}, we use the F1 score for each class and the macro-average F1 score across all classes as evaluation metrics.

\section{Experimental Results}

\subsection{Main Results}

\begin{table}[t]
\centering
\resizebox{1\linewidth}{!}{%
\renewcommand{\arraystretch}{1.2}
\begin{tabular}{lcccccc}
\toprule
    \rule{0pt}{10pt} \multirow{2}{*}{\textbf{METHOD}} & \multicolumn{6}{c}{\textbf{DEPTH}} \\ 
    \cmidrule(lr){2-7} & 1 & 2 & 3 & 4 & 5 & $\geq$ 6 \\ 
    \midrule 
     Llama3-8B & 33.42 & 40.25 & 35.91 & 36.38 & 27.19 & 35.04  \\ 
     GPT4o-mini & 34.03 & 30.94 & 37.65 & 38.83 & 38.01 & 43.59  \\ 
     GPT3.5 & 22.51 & 36.67 & 35.54 & \underline{42.84} & 39.31 & 43.54 \\ 
     Llama3-70B & \textbf{54.12} & \underline{40.90} & 39.25 & 38.69 & \underline{44.40} & 39.50 \\ 
     Qwen2.5-14B & \underline{40.23} & 39.42 & \underline{42.49} & \textbf{43.99} & 42.34 & 38.70 \\ 
    \cdashline{1-7}
     Roberta & 25.49 & 36.26 & 39.02 & 38.40 & 37.99 & \underline{46.55} \\
     Branch-BERT & 26.90 & \textbf{43.76} & 36.73 & 37.90 & 38.59 & 46.11 \\
     GLAN & 29.23 & 36.21 & 39.39 & 38.11 & 38.37 & \textbf{48.93} \\
      \textbf{SITPCL(Ours)} & 40.19 & 38.37 & \textbf{42.84} & 40.62 & \textbf{44.73} & 45.25 \\ 
    
    \bottomrule
\end{tabular}%
}
\caption{Evaluation of the performance of each model on instances with different conversation depths. The performance is reported using the F1-macro score, which encompasses the favor, against, and neutral classes.}
\label{Table 7}
\end{table}

\begin{figure*}[t]
    \centering
    \begin{subfigure}{0.3\textwidth}
        \centering
        \includegraphics[width=\textwidth]{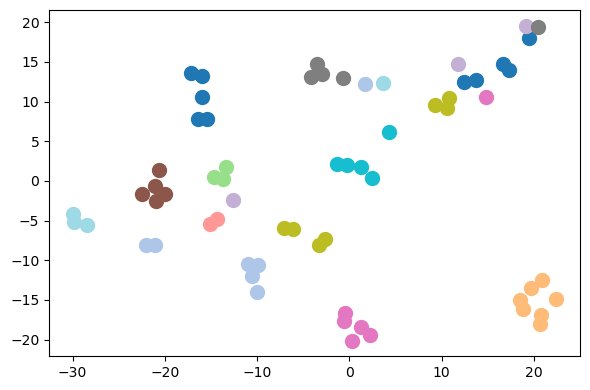}
        \caption{Branch-BERT}
        \label{fig:sub1}
    \end{subfigure}
    \begin{subfigure}{0.3\textwidth}
        \centering
        \includegraphics[width=\textwidth]{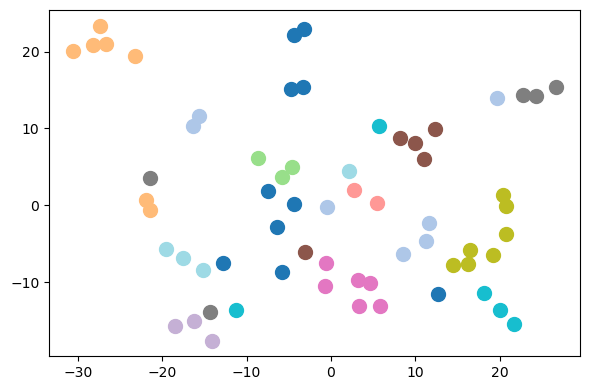}
        \caption{GLAN}
        \label{fig:sub2}
    \end{subfigure}
    \begin{subfigure}{0.3\textwidth}
        \centering
        \includegraphics[width=\textwidth]{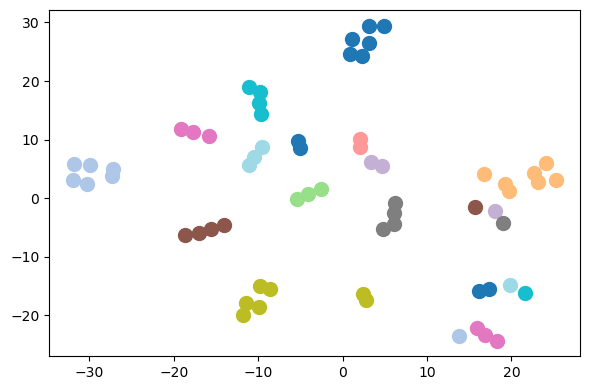}
        \caption{SITPCL}
        \label{fig:sub3}
    \end{subfigure}
    \caption{Visualization of intermediate embeddings from three models. Dots of different colors represent utterances of different targets.}
    \label{fig:5}
\end{figure*}

\begin{figure}[t]
  \centering
  \includegraphics[width=1\linewidth]{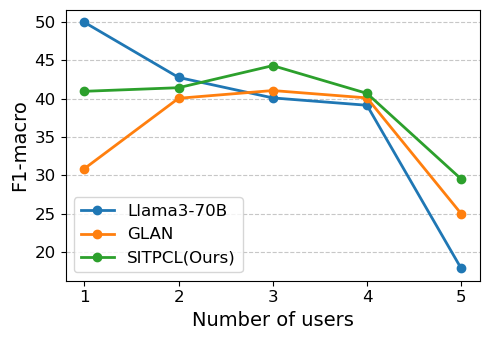}
  \caption{Evaluation of the performance of each model on instances with different user couts.}
  \label{fig:4}
\end{figure}

We present the experimental results of various models  within the target-based zero-shot conversational stance detection task, as shown in Table \ref{Table 5}. 
The mixed targets refer to the overall result on the test set without distinguishing the target type, while the noun phrase targets and the claim targets only calculate the test results of the corresponding target examples.
From the results, we make the following observations: 
1) Our proposed method, SITPCL, outperforms all baselines in the mixed-target setting, demonstrating its effectiveness.
2) The best F1-macro score for all test examples (mixed targets) is only 43.81\%, underscoring the challenge of the target-based zero-shot conversational stance detection task and dataset.
3) All methods perform significantly better on noun phrase type targets than on claim type targets, indicating the latter is more challenging in zero-shot conversational stance detection.
4) While Llama3-70B achieves the best performance on noun phrase type targets, it falls behind PLM-based fine-tuning methods on claim type targets. This suggests that while LLMs leverage extensive knowledge for noun phrases, they struggle with directly understanding claim type targets.
5) In the baseline evaluation of large language models, we observed significant performance disparities across different stance categories. For instance, in the mixed-target setting, GPT-3.5 achieved its highest performance on the neutral class with an accuracy of 53.84\%, while its performance on the favor class was considerably lower at only 28.04\%. We attribute this discrepancy to potential stance biases acquired during the model's pretraining phase or to alignment mechanisms implemented for safety, which may steer the model toward producing responses that are consistently neutral or objective.

\subsection{Ablation Study}

To rigorously assess the contribution of each module in SITPCL, we conduct ablation studies, as shown in Table \ref{Table 6}. The results indicate that removing either the SIE or TPCL module individually reduces model performance across all settings. Specifically, the F1-macro score drops by 0.90\% and 1.22\% for mixed targets when removing SIE and TPCL, respectively, demonstrating their effectiveness. Removing both modules results in a larger performance decline of 2.41\%, emphasizing that their integration enhances model optimization.

\subsection{Further Analysis}

\paragraph{Impact of Conversation Depth}

This analysis evaluates the performance of various conversational stance detection models across different conversation depths. The results are presented in Table \ref{Table 7}. Compared to other PLM-based fine-tuning methods, our model demonstrates significant superiority across all samples except those with depths of 2 and $\geq$6, confirming the effectiveness of SITPCL. Additionally, except for Llama3-8B, Llama3-70B and Qwen2.5-14B, all other methods achieve their best performance on samples with depth $\geq$6, likely because a longer conversation history provides more contextual clues for stance expression.

\paragraph{Impact of user count}
Figure \ref{fig:4} presents the evaluation results across different user counts. Our SITPCL model performs best when there are more than two users. However, as the number of users increases, all models exhibit a performance decline, highlighting the growing challenges of understanding complex user interactions.

\subsection{Error Analysis}

We randomly selected 300 error samples from the test set for detailed analysis and categorized the errors into four types:
1) Failure to understand the target (48\%): The model lacks the necessary knowledge to comprehend the target or loses track of the target during integration with the conversational context.
2) Failure to consider user interaction context (61\%): The model fails to incorporate the user's interaction history, leading to incorrect stance interpretation.
3) Over-inference (29\%): The model excessively interprets the user's stance based on limited clues, resulting in misjudgment of the user’s intent.
3) Inability to recognize sarcasm (21\%): The model struggles with identifying implicit expressions such as sarcasm, reflecting its limited pragmatic understanding.
In Appendix \ref{appendix:c}, we provide an error prediction example and model improvement analysis for each error type.

\subsection{Visualization}
To qualitatively demonstrate the SITPCL model’s ability to better capture target information, we use t-SNE to visualize intermediate utterance representations generated by different methods. As shown in Figure \ref{fig:5}, dots of the same color represent utterances associated with the same target. In both Branch-BERT and GLAN, utterance representations for different targets exhibit significant overlap. In contrast, SITPCL produces more distinct representations, with utterances related to the same target forming well-defined clusters. This validates that SITPCL’s target-aware prototypical contrastive learning effectively separates utterance representations, enhancing model performance.

\section{Conclusion}
In this paper, we present ZS-CSD, the first zero-shot multi-turn multi-party conversational stance detection dataset, and introduce a novel and challenging task: target-based zero-shot conversational stance detection. This task evaluates models on a diverse set of previously unseen targets. Compared to existing datasets, ZS-CSD contain 113 noun phrase targets and 167 claim targets. Additionally, it integrates both speaker context and speaker interaction information.
To benchmark the ZS-CSD task, we propose SITPCL, a speaker interaction and target-aware prototypical contrastive learning model. Experimental results indicate that ZS-CSD presents a significant challenge, particularly for claim type targets.

\section*{Acknowledgments}
This work is supported by the National Natural Science Foundation of China (No. 62176187).

\section*{Limitations}
Our ZS-CSD dataset is primarily derived from Chinese social media platforms and contains a limited number of targets. In the future, we aim to develop more complex zero-shot conversational stance detection datasets, incorporating multilingual and multimodal scenarios. This expansion will facilitate deeper research in the field and further advance zero-shot conversational stance detection.

\section*{Ethical Considerations}
Our dataset does not provide any personally identifiable information. Microblogs are collected using generic keywords instead of user information as queries, therefore our dataset does not have a large collection of microblogs from an individual user. Thus our dataset complies with Sina Weibo’s information privacy policy. To annotate the data, we recruited 8 senior researchers working in natural language processing, each of whom was paid \$6.5 per hour (higher than the average salary for similar jobs in the local area). The entire annotation process lasted 2 months, and the average annotation time of the 8 researchers was 100 hours.

\bibliography{custom}

\appendix

\section{Query Keywords and Domains}
\label{appendix:A}
We selected 28 keywords covering controversial topics. We grouped the 28 keywords into 6 areas: "Culture and Education", "Policy", "Science and Technology", "Healthcare", "Entertainment and Consumption" and "Social Issues" which can be seen in Table \ref{Table 2}.

\begin{table*}[t]
    \centering
    \resizebox{1\textwidth}{!}{%
    \begin{tabular}{ll}
    \toprule
    \rule{0pt}{10pt} Domain & Query Keywords \\
    \midrule
    \rule{0pt}{10pt} Cultural and Education & ACG, Chinese Traditional Culture ,
Teacher, College Entrance Examination,
Tutoring \\
    \rule{0pt}{20pt} Policy & Epidemic Prevention, Fireworks Regulations,
Family Planning, Civil Service Examination \\
    \rule{0pt}{20pt}  Science and Technology & Artificial Intelligence, Electric Vehicles, Smartphone, Autonomous Driving Technology \\
    \rule{0pt}{20pt}  Healthcare & Centralized Medical Procurement, COVID-19 Vaccine, Traditional Chinese Medicine \\
    \rule{0pt}{20pt}  Entertainment and Consumption & E-commerce, Live Streaming,
Mobile Application, Housing Prices, Star Chaser \\
    \rule{0pt}{20pt}  Social Issues  & LGBTQ, Surrogacy, XYY Syndrome, Stray Dog,
Bride-price, Elderly care, Employment \\
    \bottomrule
    \end{tabular}
    }
    \caption{The domains used in our dataset and the selected query keywords for each domain.}
    \label{Table 2}
\end{table*}

\section{Prompt for LLMs}
\label{appendix:B}

\begin{figure*}[t]
  \centering
  \includegraphics[width=0.8\linewidth]{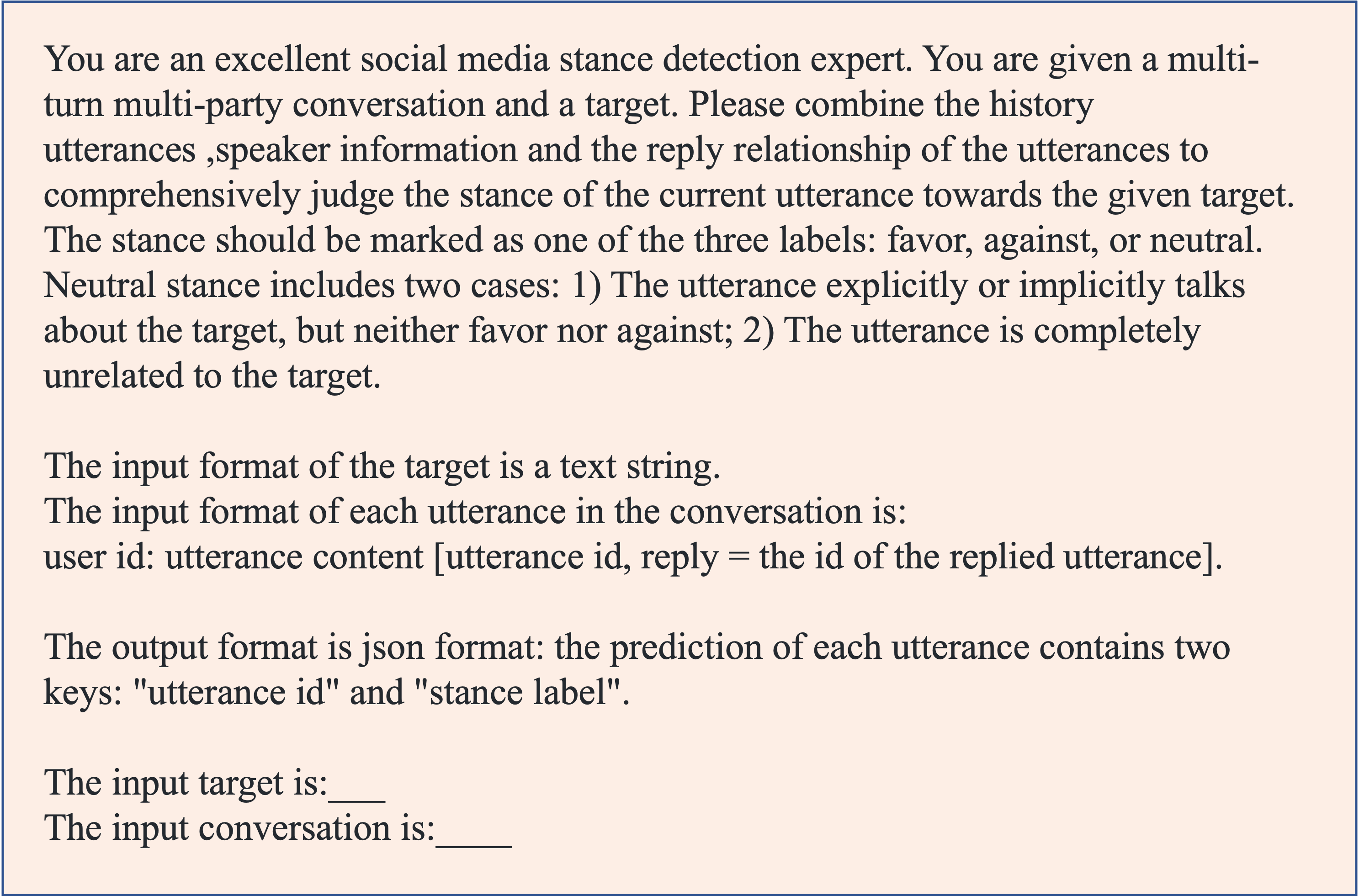}
  \caption{An example of the prompt.}
  \label{fig:3}
\end{figure*}

We use current state-of-the-art LLMs with various parameter sizes, including the closed-source GPT3.5 and open-source LLMs such as Llama3-70B and Llama3-8B. We evaluate these models exclusively under the zero-shot prompt setting.
Figure \ref{fig:3} shows the prompt we use.

\section{Error case Analysis }
\label{appendix:c}

We categorize the errors into four major types: 1) Failure to Understand the Target (the model lacks the knowledge to understand the target or loses the understanding of the target when incorporating conversation context); 2) Failure to Consider the Context of User Interaction (the model fails to take the user's interaction history into account); 3) Over-Reasoning (the model over-interprets the user's stance based on a few cues, failing to accurately grasp the user's intent); 4) Failure to Recognize Sarcasm (the model lacks the ability to understand implicit expressions such as sarcasm). Below, we provide one error prediction example and model improvement analysis for each type.

As shown in Figure \ref{ex1}, the SITPCL model did not accurately align with the target to understand User 2's stance, instead relying solely on emotional cues to classify the stance as opposing the target. In the future, the model architecture needs to be further designed to enhance its ability to be guided by and perceive the targets more effectively. As shown in Figure \ref{ex2}, the SITPCL model in this example failed to understand the context of the interaction between users. Although User 2’s last sentence alone does not clearly indicate their stance, it becomes evident when considering their previous statements. In future work, pretraining tasks could be designed to enable the model to better comprehend multi-turn conversations involving multiple participants. Additionally, the model could be improved to track each user's current stance, which could then be used to assist in future judgments when the same user appears again. As shown in Figure \ref{ex3}, from the overall context of the conversation, it is clear that User 3 is merely providing an objective analysis in response to User 2's question and is not expressing their own subjective stance. In the future, the model could be enhanced by adding auxiliary tasks for analyzing user dialogue intentions, which would help the model better understand the user's intentions and prevent it from determining the user's stance based solely on verbal cues. As shown in Figure \ref{ex4} , the SITPCL model failed to understand that User 3 was using sarcasm to oppose User 2's viewpoint, and instead interpreted it as supporting User 2's viewpoint, thus classifying it as an opposing stance. Future work should focus on enhancing the model's ability to recognize sarcasm, metaphors, and other implicit expressions.

\begin{figure*}[t]
  \centering
  \includegraphics[width=0.85\linewidth]{figs/ex1.png}
  \caption{Error of failure to understand the target.}
  \label{ex1}
\end{figure*}

\begin{figure*}[t]
  \centering
  \includegraphics[width=0.85\linewidth]{figs/ex2.png}
  \caption{Error of failure to consider the context of user interaction.}
  \label{ex2}
\end{figure*}

\begin{figure*}[t]
  \centering
  \includegraphics[width=0.85\linewidth]{figs/ex3.png}
  \caption{Error of over-reasoning.}
  \label{ex3}
\end{figure*}

\begin{figure*}[t]
  \centering
  \includegraphics[width=0.85\linewidth]{figs/ex4.png}
  \caption{Error of failure to recognize sarcasm.}
  \label{ex4}
\end{figure*}

\end{document}